\documentclass[10pt,journal,compsoc]{IEEEtran}
\usepackage{graphicx}

\usepackage{tikz}
\usepackage{comment}
\usepackage{amsmath,amssymb} 
\usepackage{color}
\usepackage{colortbl}
\usepackage{booktabs}
\usepackage{multirow}
\usepackage{wrapfig}
\usepackage{graphicx}
\usepackage{subfigure}
\usepackage{pifont}

\usepackage{url}

\usepackage{amsmath,amsfonts,bm}

\def\eqref#1{equation~\ref{#1}}

\def\1{\bm{1}}

\DeclareMathAlphabet{\mathsfit}{\encodingdefault}{\sfdefault}{m}{sl}
\SetMathAlphabet{\mathsfit}{bold}{\encodingdefault}{\sfdefault}{bx}{n}

\usepackage{xcolor}
\usepackage{orcidlink}
\usepackage{marvosym}

\usepackage{algorithm}
\usepackage{algorithmic}

\ifCLASSOPTIONcompsoc
  \usepackage[nocompress]{cite}
\else
  \usepackage{cite}
\fi

\ifCLASSINFOpdf
\else
\fi
\begin{document}
%
\title{PERF: Panoramic Neural Radiance Field from a Single Panorama}

%
%

\author{Guangcong Wang*,
        Peng Wang*,
        Zhaoxi Chen,
        Wenping Wang,
        Chen Change Loy, 
        and Ziwei Liu

\IEEEcompsocitemizethanks{\IEEEcompsocthanksitem The * denotes equal contribution. G. Wang, Z. Chen,  C. C. Loy and Z. Liu are affiliated with S-Lab, Nanyang Technological University. P. Wang is affiliated with The University of Hong Kong. W. Wang is affiliated with Texas A\&M University.
}
}

%
%

\markboth{ }%
{Shell \MakeLowercase{\textit{et al.}}: Bare Demo of IEEEtran.cls for Computer Society Journals}
%



\IEEEtitleabstractindextext{%

\begin{abstract}
Neural Radiance Field (NeRF) has achieved substantial progress in novel view synthesis given multi-view images. Recently, some works have attempted to train a NeRF from a single image with 3D priors. They mainly focus on a limited field of view with a few occlusions, which greatly limits their scalability to real-world 360-degree panoramic scenarios with large-size occlusions. In this paper, we present \textbf{PERF}, a 360-degree novel view synthesis framework that trains a panoramic neural radiance field from a single panorama. Notably, PERF allows 3D roaming in a complex scene without expensive and tedious image collection. To achieve this goal, we propose a novel collaborative RGBD inpainting method and a progressive inpainting-and-erasing method to lift up a 360-degree 2D scene to a 3D scene. Specifically, we first predict a panoramic depth map as initialization given a single panorama and reconstruct visible 3D regions with volume rendering. Then we introduce a collaborative RGBD inpainting approach into a NeRF for completing RGB images and depth maps from random views, which is derived from an RGB Stable Diffusion model and a monocular depth estimator. Finally, we introduce an inpainting-and-erasing strategy to avoid inconsistent geometry between a newly-sampled view and reference views. The two components are integrated into the learning of NeRFs in a unified optimization framework and achieve promising results. Extensive experiments on Replica and a new dataset PERF-in-the-wild demonstrate the superiority of our PERF over state-of-the-art methods. Our PERF can be widely used for real-world applications, such as panorama-to-3D, text-to-3D, and 3D scene stylization applications. Project page and code are available at \url{https://perf-project.github.io/} and \url{https://github.com/perf-project/PeRF}.
\end{abstract}


\begin{IEEEkeywords}
Panoramic Neural Radiance Field, Novel View Synthesis, 3D Scene Synthesis, Single Panorama
\end{IEEEkeywords}}

\maketitle

\IEEEdisplaynontitleabstractindextext

%
\IEEEpeerreviewmaketitle


\IEEEraisesectionheading{\section{Introduction}\label{sec:intro}}

%
%
%
%
\IEEEPARstart{N}{eural} Radiance Field (NeRF)~\cite{mildenhall2020nerf,liu2020neural,barron2021mip,barron2022mip} has recently gained increasing attention in the task of novel view synthesis for its amazing rendering quality. 
Successful training of a NeRF model generally requires multiple images from different viewpoints for supervision. However, densely capturing multiple images of a single scene requires a significant amount of effort.
In many real-world scenarios, capturing the scene with only a single image is the more convenient option for most people.
In this paper, we are interested in training a panoramic neural radiance field given a single 360-degree 2D image as input.
This promising task enables us to roam an indoor room without exhausting image capturing, which has many practical applications such as house touring, virtual-reality games, and teleconferences.



Training a panoramic neural radiance field from a single panorama has several critical challenges. \textbf{First}, a panoramic image has a 360-degree view captured at a position but does not contain 3D information. Without any 3D priors, it is impossible to train a valid NeRF from a single panorama.
\textbf{Second}, since a single panoramic image is only a partial observation of a scene due to occlusions, training a single-view panoramic neural radiance field becomes a complex coupled problem of both 3D scene reconstruction and 3D scene generation. On the one hand, we need to reconstruct visible regions of the input panorama, while on the other, we have to generate reasonable contents in invisible regions to semantically match the visible regions in a 3D space, which is difficult. \textbf{Third}, different from the 3D reconstruction of a limited field-of-view scene~\cite{xu2022sinnerf, yu2022monosdf, yu2021pixelnerf} or a 360-degree object-centric scene \cite{xu2023neurallift}, a panoramic scene often contains large-size occlusions and various open-world objects.   
\textbf{Fourth}, it is a challenge to avoid geometry conflict between visible and invisible regions. When completing the shape of invisible regions, the newly synthesized 3D geometry should not occlude the original visible regions from the given single view. Otherwise, some visible points in the given panorama will become ``invisible points" from the given view, which leads to conflicted geometry during training.

Existing methods only offer partial solutions to these problems. Instead of using dense views to train a NeRF, some NeRF methods \cite{niemeyer2022regnerf,wang2023sparsenerf,kim2022infonerf,deng2022depth,chen2021mvsnerf,jain2021putting} reduce the requirement of dense views and conduct 3D reconstruction with a few images by considering geometry constraints and high-level semantic constraints. These NeRF approaches mainly study 3D scene reconstruction from several image views but do not consider 3D scene generation, which fails to tackle invisible occlusions. Recently, a few methods \cite{xu2022sinnerf,yu2021pixelnerf} attempt to train a NeRF from a single image. They either directly use adversarial training to synthesize small-size invisible regions or need to pre-train the NeRF on other similar scenes. However, these approaches investigate limited field-of-view 3D generation from a single image. They fail to achieve good results on challenging 360-degree panoramic scenes due to large-size occlusions in a panoramic scene. Other methods \cite{li2021mine,shih20203d} proposed to use multi-plane images to represent 3D scenes instead of NeRFs. In these methods, the field of view is also limited.
Different from these previous arts, our method aims at training a panoramic neural radiance field from a single panorama, allowing 3D roaming by generating photo-realistic novel views of complex scenes. Recently, 
some methods have made initial attempts to tailor the learning of the panoramic neural radiance field from a single panorama. They introduced semantic-aware constraints \cite{kulkarni2022360fusionnerf,xu2021layout}, data augmentation of incompleted panoramas \cite{hsu2021moving}, or selection of completed images with less overlap \cite{hara2022enhancement}. These methods still fail to generate high-quality 3D panoramic neural radiance fields. Instead of studying general 360-degree scenes, some methods \cite{xu2023neurallift} study a 360-degree object-centric scene, where multiple images are captured around an object and there are few occlusions. Our goal is to generate a 3D scene from a single panorama that includes large occluded regions, which is different. Other concurrent methods \cite{weder2023removing,mirzaei2023spin} inpainted the holes for 3D, but they focus on limited field-of-view or mesh \cite{wei2023clutter}.


To address these challenges, we propose PERF, a 360-degree novel view synthesis method that trains a panoramic neural radiance field from a single panorama. 
Specifically, we predict a panoramic depth map as the initial geometry given a single panorama. We propose a collaborative RGBD inpainting method to complete RGB images and depth maps of visible regions with a trained Stable Diffusion for RGB inpainting and a trained monocular depth estimator for depth completion, aiming to generate novel appearances and 3D shapes that are invisible from the input panorama. More precisely, we warp the panoramic scene to an unseen view and complete invisible regions of interest (ROI) with a Stable Diffusion model~\cite{rombach2021highresolution}. We then predict the depth map of ROI and optimize the depth map to be consistent with the global panoramic scene (Challenges 1 \& 2). Moreover, we also propose a progressive inpainting-and-erasing method to avoid inconsistent geometry among different views (Challenges 3 \& 4). The inpainting-and-erasing method inpaints invisible regions from a random view and erases conflicted geometry regions observed from other reference views, yielding better 3D scene completion. We do this by progressively sampling novel views. We conduct experiments on the Replica and PERF-in-the-wild datasets, showing the superiority and effectiveness of our proposed PERF method.

Overall, the main contributions of this paper are \textbf{1)} PERF, a new method that trains a panoramic neural radiance field from a single panorama. To this end, we propose a collaborative RGBD inpainting with a trained Stable Diffusion model for RGB inpainting and a trained monocular depth estimator for geometry completion. Notably, the collaborative RGBD inpainting requires no extra training. 
\textbf{2)} We propose a progressive inpainting-and-erasing method to avoid geometry conflict among different views. We inpaint invisible regions by progressively increasing a random view, and erase the conflicted regions by comparing the newly added views and the reference views.
\textbf{3)} Extensive experiments on the Replica and PERF-in-the-wild datasets demonstrate that PERF achieves a new state-of-the-art in single-view panoramic neural radiance field. The proposed PERF can be applied to panorama-to-3D, text-to-3D, and 3D scene stylization tasks, showing surprising results on several promising applications.
\section{Related Work}
\label{sec:review}

\textbf{Panoramic Neural Radiance Fields.}
NeRF~\cite{mildenhall2020nerf,barron2021mip} has achieved impressive results in novel view synthesis thanks to the good representation of neural networks. Recent research extends the vanilla NeRF to large-scale scenes \cite{tancik2022block,xiangli2021citynerf,turki2022mega}, NeRF with imperfect camera poses \cite{wang2021nerf, meng2021gnerf, lin2021barf}, efficient representations \cite{Fridovich-Keil_2022_CVPR,yu2021plenoctrees,takikawa2021neural,lindell2021autoint,wizadwongsa2021nex,garbin2021fastnerf,reiser2021kilonerf,hedman2021baking,neff2021donerf,muller2022instant}, and dynamic scenes \cite{pumarola2021d,li2021neural,xian2021space,gafni2021dynamic}. Some methods focus on novel view synthesis with panoramic images \cite{gu2022omni,huang2022360roam}, omnidirectional videos \cite{gera2022casual} or object-centric projective images \cite{barron2022mip}. For example, 
Omni-NeRF \cite{gu2022omni} proposed a novel end-to-end framework for
training Neural Radiance Field (NeRF) models given only
360$^\circ$ RGB images and their rough poses. Omni-NeRF jointly learns the scene geometry and optimizes the camera parameters without knowing the fish eye projection.
360Roam \cite{huang2022360roam} proposed to learn an omnidirectional neural radiance field and progressively estimate a 3D probabilistic occupancy map to accelerate volume rendering. 
Mip-NeRF360 \cite{barron2022mip} used a non-linear scene parameterization, online distillation, and a novel distortion-based regularizer to overcome the challenges presented by unbounded panoramic scenes. 
PanoHDR-NeRF \cite{gera2022casual} proposed to capture a low dynamic range (LDR) omnidirectional video of the scene by freely waving an off-the-shelf camera around the scene and uplift the captured LDR frames to HDR. Different from these approaches that train panoramic NeRFs with multiple-view panoramas or limited field-of-view omnidirectional images, we aim at using a single panorama to train a panoramic NeRF.


\textbf{Single-view Novel View Synthesis.}
Instead of using dense views to train a NeRF, some few-shot NeRF methods~\cite{niemeyer2022regnerf,wang2023sparsenerf,kim2022infonerf,deng2022depth,chen2021mvsnerf,jain2021putting} reduce the requirement of dense views to train a NeRF by integrating 3D priors. These NeRF approaches mainly aim at 3D scene reconstruction from several image views, which cannot be applied to single-view scene generation. Recently, a few methods \cite{xu2022sinnerf,yu2021pixelnerf} attempt to train a NeRF from a single image. For example, SinNeRF \cite{xu2022sinnerf} 
used image warping to obtain geometry from unseen poses, and utilized adversarial training and high-level semantics extracted from pre-trained ViT for 3D scene completion. However, it mainly focuses on a limited field of view. When applied to a panoramic scene that contains large-size occlusions, SinNeRF fails to complete such scenes with adversarial training. PixelNeRF \cite{yu2021pixelnerf} first pre-trained a NeRF on other similar scenes and fine-tuned it on the target scene. PixelNeRF heavily depends on the knowledge transfer from similar scenes, so as to achieve promising single-view training of a NeRF. Moreover, PixelNeRF focuses on simple object-level scenes with a limited field of view. It significantly degrades when applied to complex panoramic scenes, as analyzed in \cite{xu2022sinnerf}. Some methods \cite{li2021mine,shih20203d} exploited multi-plane images to represent 3D scenes instead of NeRFs. These approaches investigate limited field-of-view 3D generation from a single image. They cannot achieve impressive results on challenging 360-degree panoramic scenes due to large-size occlusions.

To extend the limited neural radiance field to the panoramic radiance field with large-size occlusion, some methods make initial attempts to solve this problem. OmniNeRF \cite{hsu2021moving} generates new training panoramas with new virtual camera poses with a panorama and an auxiliary depth map. It directly uses visible pixels of the augmented images. 360FusionNeRF \cite{kulkarni2022360fusionnerf} introduced a semantic constraint via CLIP-ViT \cite{radford2021learning} to encourage the embeddings of different views to be as similar as possible. 360FusionNeRF also adopted a geometric constraint based on the reference depth map and warped depth maps. OmniNeRF and 360FusionNeRF cannot handle large-size occlusions. In addition, Hara et al. \cite{hara2022enhancement} proposed a method to learn a panoramic NeRF by selecting a subset of completed images that cover the target scene with less overlap of completed regions, but it is hard to preserve semantic coherence among completed images. Some methods \cite{xu2021layout} used a boundary map and a corner map as layout conditions to synthesize novel panoramic views. It does not use 3D representations and is thus inconsistent when roaming in the 3D space.

\begin{figure*}[t]
  \centering
  \includegraphics[width=0.95\linewidth]{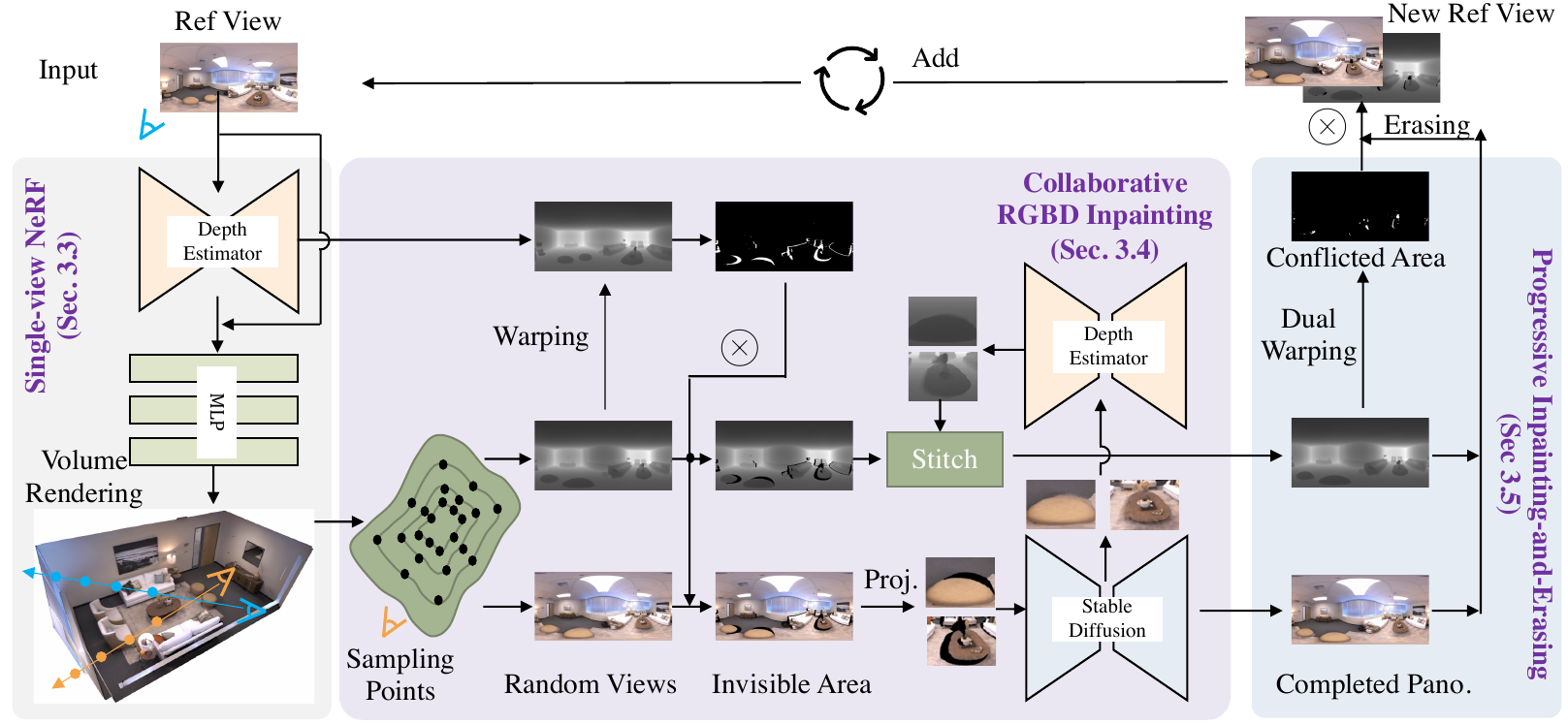}
  \caption{\textbf{Framework Overview.} PERF mainly consists of three components, including \textbf{1)} single-view NeRF training with depth maps; \textbf{2)} collaborative RGBD inpainting; and \textbf{3)} progressive inpainting-and-erasing. 
  Specifically, given a single panorama, we predict its depth map with a Depth Estimation model and train a NeRF with the input view as initialization. Then a collaborative RGBD inpainting module that contains a Depth Estimator and a Stable Diffusion is proposed to extend NeRF to random views. To avoid geometry conflict, a progressive inpainting-and-erasing module is used to compute a mask of the conflicted regions and eliminate these regions. We fine-tune the panoramic NeRF with the reference single-view panorama and new panoramas generated from random viewpoints until convergence.}
  \label{fig:framework}
\end{figure*}

In addition, some methods \cite{weder2023removing,mirzaei2023spin} inpainted the holes for 3D, but they focus on a limited field-of-view and do not fully address geometry consistency. For example, SPIn-NeRF inpaints depth maps and uses a perceptual loss to complete the RGB appearance of a scene, and does not introduce mechanisms to avoid geometry consistency. Weder et al. \cite{weder2023removing} proposed to re-weight multiple views to reduce geometry inconsistency based on confidence scores of views. A concurrent work \cite{wei2023clutter} introduced RGB inpainting, image-guided depth completion, and view-consistency inspired refinement for mesh reconstruction. Compared with \cite{wei2023clutter}, our method is different in several aspects: 1) we focus on single-view panoramic NeRF while it focuses on removing objects and inpainting meshes; 2) we design trajectories and sample views for reconstruction and it performs clutter segmentation; 3) we directly predict depth maps of regions of interest and stitch depth maps while it uses image-guided depth completion. OtherSLAM methods \cite{zhu2022nice,sucar2021imap} aim at simultaneous localization and mapping, which 
takes an RGB-D image stream as input and outputs a camera pose and a scene representation. They significantly differ from the proposed method in that our method only requires a single panorama.


\section{Our Approach}
Training a panoramic neural radiance field (NeRF) from a single panorama is a challenging problem due to \textbf{1)} lack of 3D information, \textbf{2)} large-size object occlusion, \textbf{3)} a coupled problem on reconstruction and generation, and \textbf{4)} geometry conflict between visible regions and invisible regions during inpainting. Some methods resort to direct interpolation \cite{hsu2021moving} of NeRFs, adversarial learning \cite{xu2022sinnerf}, high-level semantic constraints \cite{xu2022sinnerf,jain2021putting,kulkarni2022360fusionnerf}, or pre-training on similar scenes \cite{yu2021pixelnerf} to recover invisible regions. However, in complex panoramic scenes, large-size occlusions often occur when arbitrary human viewpoints are allowed. Directly using these approaches to solve single-view panoramic NeRF struggles to synthesize high-quality 3D semantics and geometry. In this paper, we present PERF to tackle these challenges. In Section \ref{subsec:prelimiary}, we give a brief introduction to NeRF and problem definition. Then, we provide an overview of PERF in Section \ref{subsec:overview}. Specifically, we train a single-view NeRF with depth maps for visible regions of a scene in Section \ref{subsec:single}. We integrate a collaborative RGBD inpainting method to guide the training of a NeRF to complete RGB images and depth maps of regions of interest as presented in Section \ref{subsec:dual}. To avoid geometry conflict between visible regions and invisible regions, we propose a progressive inpainting-and-erasing scheme in Section \ref{subsec:dual_warp}. Finally, we provide an overview of the whole PERF algorithm in Section \ref{sec:algo}.

\subsection{Preliminary and Problem Definition}
\label{subsec:prelimiary}
\noindent\textbf{Neural Radiance Fields.} The goal of a NeRF \cite{mildenhall2020nerf} is to learn a neural function $f$ that maps a 3D spatial location $\mathbf{x}$ and a viewing direction $\mathbf{d}$ into a volume density $\sigma$ and a color value $\mathbf{c}$.
The expected pixel color $\hat{C}(\mathbf{r})$ is rendered by casting a ray $\mathbf{r}(t)=\mathbf{o}+t\mathbf{d}$ with near and far bounds $t_n$ and $t_f$, which is partitioned into $N$ points ($t_1, t_2, ..., t_N$) along a ray $\mathbf{r}$. The color $\hat{C}(\mathbf{r})$ is blended by $\hat{C}(\mathbf{r})=\sum_{i=1}^{N}w_i\mathbf{c}_i$, where $\quad w_i=T_i(1-\mathrm{exp}(-\sigma_i\delta_i))$,  $T_i=\mathrm{exp}(-\sum_{j=1}^{i-1}\sigma_j\delta_j)$ and $\delta_i=t_{i}-t_{i-1}$. The NeRF is trained by a reconstruction loss, which is given by
\begin{equation}
\label{eq:reconstr}
\mathcal{L}_{\mathrm{nerf}}=\sum_{\mathbf{r}\in R}||\hat{C}(\mathbf{r})-C(\mathbf{r})||^{2},
\end{equation}
where $C(\mathbf{r})$ is the ground-truth pixel color. 


\noindent\textbf{Problem Definition.} Conventional NeRFs aim to learn a mapping function ${f}$ with images from multiple views by minimizing the color reconstruction loss in \textbf{Eq.~(\ref{eq:reconstr})}.
When the number of views is reduced to one, the optimization problem is under-constrained. The goal of this paper is to train a panoramic NeRF to recover a 3D scene from a single RGB panorama.

\subsection{Overview of PERF} 
\label{subsec:overview}
The framework of PERF is illustrated in Figure \ref{fig:framework}. PERF mainly consists of three components, including \textbf{1)} single-view NeRF training with depth supervision; \textbf{2)} collaborative RGBD inpainting of ROI (region of interest); and \textbf{3)} progressive inpainting-and-erasing generation.
Specifically, given a single RGB panorama, we predict its depth map with a depth estimation model~\cite{gu2022omni} and train a NeRF with both the RGB panorama and the depth map as initialization.
Then, starting with the input panorama as a reference view, we perform a collaborative RGBD inpainting process to extend NeRF to several random views. For each randomly sampled pose, we use Stable Diffusion~\cite{rombach2021highresolution} to complete the appearance of occluded regions and then complete the shape of these regions by depth prediction and scale-invariant stitching. We note that the newly completed geometry of the invisible region may have conflicts with the observations of the already-learned reference views. To this end, we propose a progressive inpainting-and-erasing strategy to compute a mask of the conflicted regions and eliminate these regions from the NeRF training.
We then fine-tune the panoramic NeRF with the reference panorama views and the new panorama generated from the random viewpoint. After fine-tuning, the new panorama is also set as a reference panorama for the upcoming NeRF learning of the random viewpoint. We progressively enlarge the range of viewpoints of the NeRF by randomly sampling a camera pose and training the NeRF until convergence. With these three components, we achieve a photo-realistic panoramic NeRF with a single-view panorama, enabling users to roam in a complex scene.


\subsection{Single-view NeRF Training with Depth Map}
\label{subsec:single}

To start with, we initialize the geometry of the visible regions with a 360-degree depth estimation method. To this end, we use a similar approach following 360MonoDepth~\cite{rey2022360monodepth} that predicts the depth map of the panorama by fusing multiple depth maps of perspective-projected images. 
Specifically, given a single panorama $I_{\mathrm{rgb}}$, we project the image to $K$ images in perspective projections $\{ \bar{I}_{\rm rgb}^{k} \}_{k=1}^K$ by different directions. We set $K=20$. We do this because the existing monocular depth estimator trained on large-scale datasets only works for limited field-of-view images instead of panoramas that warped 360-degree scenes into a rectangular matrix.
Then we use the depth estimation model~\cite{eftekhar2021omnidata} to estimate the depth values of $\{\bar{I}_{\rm rgb}^{k} \}$ and obtain the depth map $\{\bar{I}_{\rm d}^{k} \}$.
Note that due to the scale ambiguity of single-view depth estimation, the estimated depth values of those overlapped regions can be largely different. 
Therefore, after estimating the perspective depth maps, we perform the deformable align operation to jointly align those depth maps to a canonical learnable panorama depth map $I_{\mathrm{d}}$.
For each perspective depth map $\{\bar{I}_{\rm d}^{k} \}$, we assign it a scale value $w_{k}$ and local bias values $\{ b^k_{i,j} \}$. To perform the alignment, we optimize
\begin{equation}
\label{eq:align}
\sum_{k,i,j}||\bar{v}^k_{i,j}w+b^k_{ij}-v_{\mathcal{M}({i,j})}||+\mathcal{L}_{s}(b_{i,j+1},b_{i,j})+\mathcal{L}_{s}(b_{i+1,j},b_{i,j}),
\end{equation}

Eq. (\ref{eq:align}) is composed of a data alignment term and a total variation term~\cite{rudin1994total}. Here $\bar{v}^k$ denotes the pixel value of depth map $\{\bar{I}_{\rm d}^{k} \}$, $v$ denotes the pixel value of a panorama depth map $I_{\mathrm{d}}$, $\mathcal{L}_s$ is the smooth L1 loss, and $\mathcal{M}$ is a projective operation that maps the perspective image coordinates to panorama image coordinates.
With both RGB panoramas and depth maps, we train a single-view NeRF for visible regions. Specifically, given a single panorama $I_{\mathrm{rgb}}$, we predict its depth map as $I_{\mathrm{d}}$ by the depth estimator $f_{\mathrm{d}}$.
We integrate $I_{\mathrm{d}}$ into the training of a NeRF, which is given by
\begin{equation}
\label{eq:reconstr_depth}
\mathcal{L}_{\mathrm{nerf}}^{'}=\sum_{\mathbf{r}\in R}(||\hat{C}(\mathbf{r})-{C}(\mathbf{r})||^{2}+||\hat{D}(\mathbf{r})-{\bar{D}}(\mathbf{r})||^{2}),
\end{equation}
where $\hat{D}(\mathbf{r})$ denotes the expected depth values rendered by the NeRF and ${\bar{D}}(\mathbf{r})$ denotes the corresponding depth pixels computed by the depth estimator. We optimize $\mathcal{L}_{\mathrm{nerf}}^{'}$ to train a NeRF to be aware of the 3D shape of the visible regions. Since only a single panorama is available, the trained NeRF $f_{\mathrm{nerf}}$ only works for visible regions. We then complete the invisible regions of the scene as described below.  

\subsection{Collaborative RGBD Inpainting with Stable Diffusion and Monocular Depth Estimator}
\label{subsec:dual}
Single-view NeRF training with depth maps provides accurate reconstruction for visible regions of a scene. For the occluded regions, the trained NeRF yields blurred interpolation because we train the NeRF with pixel-wise reconstruction only for visible regions. Pixel-wise reconstruction does not consider high-level semantics and is thus hard to interpolate large-size occluded regions. Therefore, after learning the geometry and appearance of the visible region from the input view, we need to guide the NeRF model to infer the invisible regions in a 3D space, given our learned regions as contextual semantic information. On the one hand, we have the freedom to choose paths for inpainting in a 3D scene; on the other hand, we have to complete invisible 3D regions given invisible 3D regions. We address the inpainting trajectory problem by sampling points according to the estimated single-view panoramic depth map.


More precisely, given our estimated single-view panoramic depth map, we get the distance values of all pixels on zero elevation angle position.
These distance values form a closed curve on the horizontal plane. Noting that the distance values can change suddenly between neighboring pixels, we apply 1D Gaussian filtering on the distances and get a smoothed closed curve $\mathcal{C}$.
This curve defines the maximum space within which the camera can move.
To sample new camera positions, we progressively sample camera positions on shrunk curves $\{ \mathcal{C}_i \}$ whose scale is $\{s_i\}$ times the largest curve $C$.
In our experiments, the number of shrunk curves is $4$ and the scale factors $\{s_i\}$ are $0.15$, $0.3$, $0.45$, and $0.6$ respectively.
We uniformly sample $8$ camera positions on each shrunk curve $\{s_i\}$. As a result,  $32$ positions are sampled in total.

Then we address the RGBD inpainting problem in a 3D scene. The intuitive idea is directly training an RGBD inpainting model on a large-scale RGBD dataset. Currently, the largest RGBD dataset is MIX 6 described in DPT \cite{ranftl2020towards}, which combines multiple diverse public datasets and contains about 1.4 million images. However, the state-of-the-art text-to-image generation model conditioned on language prompts, Stable Diffusion, is trained on LAION-400M \cite{schuhmann2021laion}, which contains CLIP-filtered 400 million image-text pairs. An inpainting version of Stable Diffusion resumed from the base Stable Diffusion, adds input channels for masked images, and is further fine-tuned for RGB inpainting. It motivates us to make full use of such a large RGB inpainting model pre-trained on a 400M-level dataset to help RGBD inpainting. Our idea is to perform collaborative RGBD inpainting by marrying the Stable Diffusion with the monocular depth estimator DPT \cite{ranftl2020towards}.

To achieve this goal, we have to 1) identify the invisible regions and get the projective 2D masks and RGB images (limited field-of-view images) of occluded regions for each newly sampled camera view; 2) inpaint RGB images with an off-the-shelf Stable Diffusion model and use a depth estimator to collaboratively inpaint the corresponding depth maps for a newly sampled camera view. In this way, the appearance and geometry of these occluded regions are completed.




\textbf{Identify Invisible Regions.} In Section \ref{subsec:single}, we train a single-view NeRF which provides accurate reconstruction for visible regions of a 3D scene but fails to interpolate photo-realistic invisible regions. After a single-view NeRF is trained, we cannot directly identify if a 3D point is visible from a given reference view. 

\begin{figure}[t]
  \centering
  \includegraphics[width=0.95\linewidth]{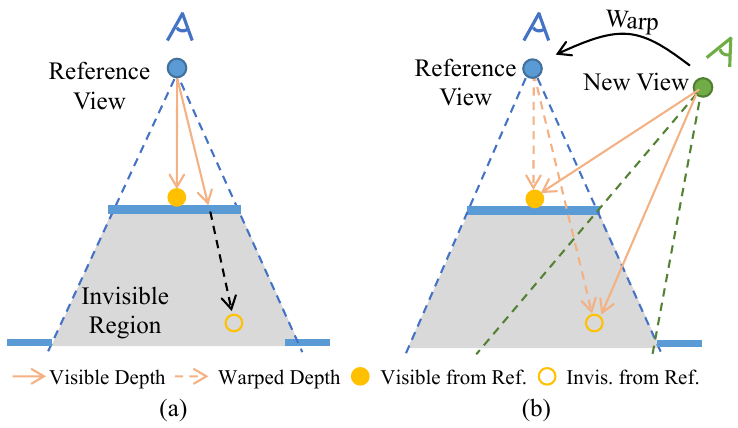}
  \vspace{0pt}
  \caption{Identify invisible regions. (a) The depth values from a given reference view. (b) The depth values warped from a new view to the reference view. For example, the yellow solid point is warped from a new view to the reference view, and the depth values are the same in (a) and (b). Therefore, it is visible. When the yellow hollow point in (b) is warped from the new view to the reference view, the wrapped depth value is greater than the visible depth value in (a). We identify it as an invisible point (Best viewed in color). }
  \label{fig:check_invisible}
  \vspace{0pt}
\end{figure}

In this section, we propose to identify the invisible regions by a warped depth check scheme. Specifically, for a newly sampled camera pose $p^{'}$, we perform volume rendering and obtain a panoramic depth map $I_\mathrm{d}^{'}$ and an RGB panorama $I_{\mathrm{rgb}}^{'}$. To distinguish whether a pixel is visible or invisible, for each reference view (reference views will gradually increase), we warp the depth map $I_d^{'}$ back to the viewpoint $p$ of the reference panorama and distinguish if each depth value matches the corresponding depth value of $I_{d}$, which is given by
\begin{equation}
\label{eq:mask}
m_{ij}^{'}=\mathbb{I}(\mathrm{warp}(v_{ij}^{'})-v_{ij} > \varepsilon),
\end{equation}
where $v_{ij}^{'}$ and $v_{ij}$ are pixel values of $I_\mathrm{d}^{'}$ and $I_\mathrm{d}$, respectively. The $\varepsilon$ is a small value that allows a small error. The $\mathrm{warp}(\cdot)$ is a warping function. The $\mathbb{I}(\cdot)$ denotes an indicator function. The $m_{ij}^{'}$ denotes a pixel of a mask $m^{'}$, indicating invisible (1) or visible (0). As illustrated in Fig. \ref{fig:check_invisible} (a), we can compute the depth value from a reference view to the surface of a scene given a trained NeRF. In Fig. \ref{fig:check_invisible} (b), we warp two points from a new view to the reference view. After warping, it is observed that the depth of the yellow solid point remains unchanged while the warped depth of the yellow hollow point is larger than the depth in Fig. \ref{fig:check_invisible} (a). We conclude that the yellow solid point is visible and the yellow hollow point is invisible according to Eq. (\ref{eq:mask}).
The mask $m^{'}$ helps us recognize the invisible regions of each reference view to guide the upcoming inpainting process.
The final mask $m^{'}$ indicating the invisible regions is computed by performing the logic AND operation of all the masks regarding the reference views (we progressively add reference views as described in Section \ref{subsec:dual_warp}). We use $m^{'}$ to mask $I_{\mathrm{rgb}}$ and $I_{\mathrm{d}}$, and get the masked RGB image ${\bar{I}}_{\mathrm{rgb}}=I_{\mathrm{rgb}}\otimes (1-m^{'})$ and the masked depth map ${\bar{I}}_{\mathrm{d}}=I_{\mathrm{d}}\otimes (1-m^{'})$ for RGBD inpainting (see Fig. \ref{fig:framework}). 


\textbf{Collaborative RGBD Inpainting.} Then we complete the contents of these invisible regions. The existing methods, such as SinNeRF \cite{xu2022sinnerf,kulkarni2022360fusionnerf}, directly use adversarial learning or high-level semantic features to fill occlusions. However, in complex 3D panoramic scenes, large-size occlusions often occur. Directly using adversarial learning or high-level semantic constraints fails to fill the large-size invisible regions, which will be shown in Sections~\ref{subsec:exp_replica} and \ref{subsec:perf-wild}. In contrast, we propose a collaborative RGBD inpainting method that integrates a diffusion-based RGB inpainting model \footnote{\url{https://github.com/Stability-AI/stablediffusion}} and a monocular depth estimator DPT \footnote{\url{https://github.com/isl-org/DPT}}. Given masked RGB and depth panoramas ${\bar{I}}_{\mathrm{rgb}}$ and ${\bar{I}}_{\mathrm{d}}$, noting that the content organization of the panorama images are different from the usually used perspective cameras, which are out-of-domain for most main-stream pre-trained models, we first select the regions of interest (ROI) that contain partial invisible areas with bounding boxes and project them into normal FOV (field-of-view) images, denoted as ${\bar{I}}_{\mathrm{rgb}}^{\mathrm{ROI}}$ and ${\bar{I}}_{\mathrm{d}}^{\mathrm{ROI}}$. We use Stable Diffusion \cite{rombach2021highresolution} to complete ${\bar{I}}_{\mathrm{rgb}}^{\mathrm{ROI}}$ as ${\hat{I}}_{\mathrm{rgb}}^{\mathrm{ROI}}$. Here, the inpainting model combines the latent VAE representations of the masked images as an additional conditioning. It uses additional input channels for U-Net, which is zero-initialized and fine-tuned with inpainting datasets. Specifically, the Stable Diffusion model takes ${\bar{I}}_{\mathrm{rgb}}^{\mathrm{ROI}}$ and its mask as conditions, and outputs an inpainted RGB image. Instead of directly inpainting depth maps, we use a depth estimator to collaboratively complete depth maps. We infer the geometry of the newly inpainted image ${\hat{I}}_{\mathrm{rgb}}^{\mathrm{ROI}}$ by predicting the depth map ${\hat{I}}_{\mathrm{d}}^{\mathrm{ROI}}$ with a monocular depth estimator. Because the visible region of ${\bar{I}}_{\mathrm{d}}^{\mathrm{ROI}}$ is known, our idea is to align the visible region between ${\bar{I}}_{\mathrm{d}}^{\mathrm{ROI}}$ and ${\hat{I}}_{\mathrm{d}}^{\mathrm{ROI}}$, and thus inpaint the invisible geometry. Since the scale and shift are different between ${\hat{I}}_{\mathrm{d}}^{\mathrm{ROI}}$ and ${\bar{I}}_{\mathrm{d}}^{\mathrm{ROI}}$, we stitch ${\hat{I}}_{\mathrm{d}}^{\mathrm{ROI}}$ into ${\bar{I}}_{\mathrm{d}}^{ROI}$ with a linear-invariant constraint to align the visible regions and fill invisible regions, which is optimized by 
\begin{equation}
\label{eq:stiching}
\mathop{\arg \min}\limits_{w,\mathbf{b}} \sum_{i,j}||\hat{v}_{i,j}w+b_{ij}-\bar{v}_{i,j}||+\mathcal{L}_{s}(b_{i,j+1},b_{i,j})+\mathcal{L}_{s}(b_{i+1,j},b_{i,j})
\end{equation}
where $\hat{v}_{i,j}$ and $\bar{v}_{i,j}$ are pixels of ${\hat{I}}_{\mathrm{d}}^{\mathrm{ROI}}$ and ${\bar{I}}_{\mathrm{d}}^{\mathrm{ROI}}$. The $w$ denotes a scalar shared by all pixels of the ROI. The $\mathbf{b}$ is the same size as the ROI. The $\mathcal{L}_s$ is a smooth L1 loss. The first term regularizes  ${\hat{I}}_{\mathrm{d}}^{\mathrm{ROI}}$ and ${\bar{I}}_{\mathrm{d}}^{\mathrm{ROI}}$ to be linear-invariant. The second and third terms are the TV loss inspired by \cite{rudin1994total}. After aligning ${\hat{I}}_{\mathrm{d}}^{\mathrm{ROI}}$ and ${\bar{I}}_{\mathrm{d}}^{\mathrm{ROI}}$, we can stitch ${\hat{I}}_{\mathrm{d}}^{\mathrm{ROI}}$ into a masked panorama ${\bar{I}}$ and obtain a completed panorama.

The completed RGB images and depth of the panorama from a new viewpoint are then used to fine-tune the NeRF, which encourages the NeRF to generate reasonable novel views from invisible regions. We progressively perform the collaborative RGBD inpainting by sampling camera poses until convergence. However, the collaborative RGBD inpainting independently completes invisible regions for each view, leading to inconsistent 3D geometry from different views.

\begin{figure}[t]
  \centering
  \includegraphics[width=0.95\linewidth]{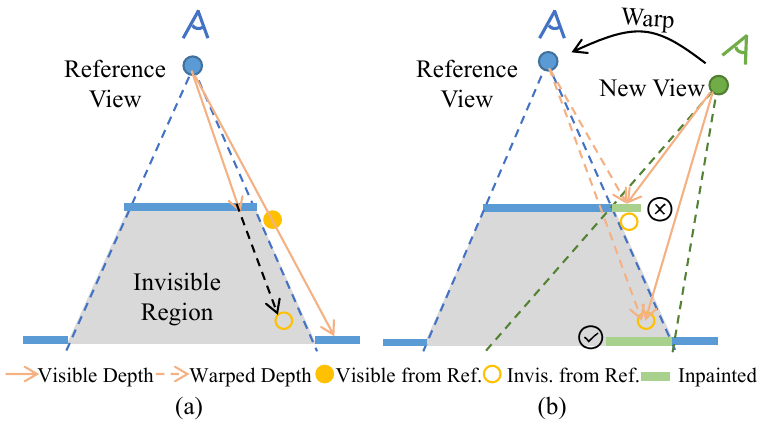}
  \vspace{0pt}
  \caption{Conflicted geometry between reference view and a random new view. (a) Before inpainting: the gray region is invisible to the reference view due to occlusion. (b) After inpainting: direct inpainting (the green segments, newly inpainted geometry) of the indivisible region from a new view might cause inter-view geometry inconsistency. The upper point is visible in (a) but invisible in (b) from the reference view. The lower point is invisible in both (a) and (b) from the reference view. Therefore, the lower inpainted segment is acceptable, while the upper inpainted segment leads to geometry conflict. We detect these conflicted regions by Eq. (\ref{eq:mask2}), i.e., the warped depth is smaller than the visible depth from the reference view (Best viewed in color).}
  \label{fig:PERF-vis-geo-confict}
  \vspace{0pt}
\end{figure}




\subsection{Progressive Inpainting-and-Erasing Generation}
\label{subsec:dual_warp}
The proposed collaborative RGBD inpainting does not guarantee consistent geometry between the reference panorama and the new panorama from a random novel view. As illustrated in Fig.~\ref{fig:PERF-vis-geo-confict}, we first train a single-view NeRF given a reference view shown in Fig.~\ref{fig:PERF-vis-geo-confict} (a). From a new view, the model inpaints two green segments (e.g., desk and floor), as shown in Fig. \ref{fig:PERF-vis-geo-confict} (b). We find that the upper point is visible in Fig.~\ref{fig:PERF-vis-geo-confict} (a) and becomes invisible from the reference view as observed in Fig.~\ref{fig:PERF-vis-geo-confict} (b). That is, the newly synthesized geometry of the invisible regions (the upper green segment in Fig.~\ref{fig:PERF-vis-geo-confict} (b)) occludes some visible regions from the reference viewpoint (e.g., the upper point). This occlusion makes a part of visible regions of reference view become invisible regions, leading to geometry conflict across different views. It hurts the training of the NeRF with conflicted views and the geometric generation of a 3D scene.

To avoid the conflicted geometry, our idea is an inpainting-and-erasing strategy. Inpainting aims at completing invisible regions and erasing aims at removing conflicted inpainted regions. We perform RGBD inpainting from a sampled view and erase conflicted regions of the new view according to reference views. The inpainted-and-erased new view is then added to the collection of reference views. Specifically, we first compute a mask $M^{''}$ that denotes the conflicted geometry between the new view and the reference views. This is achieved by a similar depth-checking strategy used in invisible region detection in Section~\ref{subsec:dual}. As shown in Fig. \ref{fig:PERF-vis-geo-confict}, for the upper inpainted segment (conflicted geometry), the warped depth value in Fig. \ref{fig:PERF-vis-geo-confict} (b) is smaller than the visible depth value in Fig. \ref{fig:PERF-vis-geo-confict} (a). For the lower inpainted segment (no conflict), the warped depth is larger than the visible depth value. Formally, considering checking the geometry consistency of the new view with the inpainted depth map, for each learned reference view, we warp the depth map back to the reference panorama and check if the depth value is smaller than the corresponding depth value of $I_{d}$, which is given by
\begin{equation}
\label{eq:mask2}
m_{ij}^{''}=\mathbb{I}(v_{ij}-\mathrm{warp}(v_{ij}^{'}) > \varepsilon ),
\end{equation}
where $v_{ij}^{'}$ and $v_{ij}$ are pixel values of $I_d^{'}$ and $I_d$, respectively. The $m_{ij}^{''}$ denotes a pixel of a mask $m^{''}$, indicating inconsistent (1) or consistent (0) geometry. 
The final mask of the conflict regions is computed by the logic AND operation of all the individual mask of the reference views. After the conflict regions are identified, we eliminate these regions from the supervision of NeRF. We iterate inpainting invisible regions and eliminating conflicted regions from sampled novel views until the algorithm converges to a completed 3D geometry. Note that the first warping operation in Section~\ref{subsec:dual} is to detect invisible regions and the second warping operation in this section is to detect the conflicted inpainted regions that violate the consistent geometry constraint.



\subsection{Algorithm}
\label{sec:algo}
In this section, we provide an overall pseudo algorithm of PERF to make it clear, as shown in Algorithm \ref{alg:alg}. Basically, given a single panorama $I_{\mathrm{rgb}}$, we first estimate its depth map as $I_{\mathrm{d}}$. Given $I_{\mathrm{rgb}}$ and $I_{\mathrm{d}}$, we train a panoramic NeRF with a single-view panorama. Training a single-view NeRF only works for visible regions but fails to interpolate geometry and appearance for the invisible regions. To address this problem, we start with sampling novel views based on the depth map $I_{\mathrm{d}}$. Gvien a sampled novel view, we compute the invisible regions and get an invisible mask according to Eq. (\ref{eq:mask}). We perform collaborative RGBD inpainting with a Stable Diffusion and a pre-trained monocular depth estimator DPT. Since the RGBD inpainting is performed for each view independently, it leads to consistent geometry from different views. To solve this problem, we compare the sampled view and each reference view, and erase the conflicted regions. The inpainted-and-erased new view is added into the reference view set. We fine-tune the panoramic NeRF with all of the reference views.

\begin{algorithm}
\caption{Algorithm of PERF}
\label{alg:alg}
 \begin{algorithmic}[1]
 \renewcommand{\algorithmicrequire}{\textbf{Input:}}
 \renewcommand{\algorithmicensure}{\textbf{Output:}}
 \REQUIRE A single panorama $I_{\mathrm{rgb}}$
 \ENSURE  A panoramic NeRF $f_{\mathrm{nerf}}$
  \STATE Estimate a depth map $I_{\mathrm{d}}$ of the given panorama $I_{\mathrm{rgb}}$ based on Eq. (\ref{eq:align}), and obtain $I_{\mathrm{rgbd}}$.
   \STATE Train a single-view NeRF $f_{\mathrm{nerf}}$ with $I_{\mathrm{d}}$ and $I_{\mathrm{rgb}}$ based on Eq. (\ref{eq:reconstr_depth}).
   \STATE Generate trajectories $\{p_k\}$ of novel views based on the depth map.
   \STATE Initialize a set of reference views as $\mathbf{I}_{\mathrm{ref}}=\{I_{\mathrm{rgbd}}\}$
  \FOR {$n \gets 1$ to $iters$}
  \STATE Randomly sample a new camera view $p^{'}$ from $\{p_k\}$.
  \STATE Compute the invisible regions according to Eq. {\ref{eq:mask}}.
  \STATE Perform collaborative RGBD inpainting using a Stable Diffusion for RGB inpainting and a depth estimator for stitching (Eq. \ref{eq:stiching}).
  \FOR {$\forall I_{\mathrm{ref}}\in \mathbf{I}_{\mathrm{ref}}$}
   \STATE Compute the conflicted region and erase it  according to Eq. (\ref{eq:mask2}).
  \ENDFOR
  \STATE Add this new view into the reference views $\mathbf{I}_{\mathrm{ref}}$.
  \STATE Finetune $f_{\mathrm{nerf}}$ with $\mathbf{I}_{\mathrm{ref}}$.
  \ENDFOR

 \end{algorithmic} 
\end{algorithm}


\section{Experiments}
\noindent\textbf{Datasets.} We conduct experiments on the Replica~\cite{replica19arxiv} dataset for both quantitative and qualitative evaluations. The Replica dataset contains eight scenes. On the Replica dataset, for each scene, we render three panorama images with the resolution of $2048\times 1024$, where one is used for training and the others for testing. The rendered testing images are not from the sampled views in the training. To demonstrate the generalization of PERF, we also evaluate the proposed method on in-the-wild data for qualitative evaluation. We collect a new dataset, PERF-in-the-wild, which contains eight indoor scenes downloaded from the Internet. Each scene has only one single panorama. 

\noindent\textbf{Evaluation Metrics and Protocols.}
We adopt four evaluation metrics in our experiment.
Among them, three metrics are the standard metrics of view synthesis computed against the whole testing images - PSNR, SSIM~\cite{wang2004image}, and LPIPS~\cite{zhang2018unreasonable}. In addition to these three metrics, we report a ``masked'' version of PSNR. In this metric calculation, for each testing view, we generate the mask of invisible regions by our aforementioned collaborative RGBD inpainting and progressive inpainting-and-erasing methods, and only the pixels that are in the invisible regions are counted for the evaluation of PSNR. The masked PSNR is a key metric to measure the single-view panoramic 3D generation since synthesizing invisible regions is the most challenging problem.


To evaluate pixel-level metrics, we need to align the rendered images and ground-truth images. During the training of PERF, the 3D geometry of PERF is initialized by a depth estimator and followed by geometry inference, where there is a scale ambiguity between our predicted geometry and the ground-truth geometry. Therefore, to correctly evaluate the rendered images from novel views, we conduct a global scale alignment between the predicted depth map of the reference view and the ground-truth depth map.

\noindent\textbf{Implementation Details.}
We use Instant-NGP \cite{muller2022instant} for the representation of the NeRF model. The hash table size is $2^{19}$ with $16$ levels of feature vectors of channel size $2$.
To learn PERF, we first train NeRF with the single input panorama in 10k iterations. We then progressively sample $32$ new positions with approximately gradually increased distance from the original position of the reference panorama view. For each newly added reference view, after our collaborative RGBD inpainting and progressive inpainting-and-erasing, we continue the training of NeRF for 2000 iterations, where for the first 100 iterations we only use the new reference view for supervision, and for the other 1900 iterations we supervise NeRF with all the reference views.
We use Adam~\cite{kingma2014adam} to optimize the parameters of the NeRF. For each learning stage of a newly sampled view, the learning rate is initially set to $10^{-3}$ and then decayed to $10^{-4}$ with the cosine annealing strategy. During training at each iteration, we randomly sample 2048 rays with ground-truth or inpainted colors and depth values to supervise NeRF. Training a scene usually takes about 2 hours on a single Nvidia V100 GPU with 32G memory.

More precisely, we use NeRFAcc~\cite{li2022nerfacc} framework to implement the Instant-NGP~\cite{muller2022instant} NeRF training and rendering. We use the pre-trained model from Stable-Diffusion-2~\cite{stability2022} as our inpainting model.
We note that the training images of the inpainting model are mostly captured in perspective field-of-view camera projections and it is not proper to directly use the inpainting model on the panoramas, where distortions exist.
Therefore for all the masked regions, we project the regions to the perspective views, adopt inpainting, and then project back the inpainted region to the panoramas. To regularize NeRF from foggy effects, for PERF and all the baseline methods that are integrated into our panorama framework, we additionally adopt the distortion loss~\cite{barron2022mip} with weight $\lambda_{dist}=1$ to encourage the volume densities of NeRF to converge to a valid surface instead of ``floating'' everywhere.


\noindent\textbf{Comparison Methods.}
To demonstrate the effectiveness of our proposed method, we compare PERF with the following three state-of-the-art few-shot NeRF techniques and two recent single panorama view synthesis works.

{\bf 1)} DS-NeRF~\cite{deng2022depth} increases the generalization ability of vanilla NeRF on few-shot novel view synthesis by adding a depth constraint for geometry regularization. However, it cannot synthesize occluded regions well.

{\bf 2)} DietNeRF~\cite{jain2021putting} regularizes the rendered novel prospectively projected views of NeRF by adding a semantic regularization, which is achieved via computing the cosine similarity between novel views and reference views of CLIP~\cite{radford2021learning} features.
To integrate DietNeRF in our framework or panoramic NeRF learning, we randomly render prospectively projected image patches with NeRF on the occluded regions of the novel views and optimize the CLIP~\cite{radford2021learning} loss against the image patches of the reference input view.

{\bf 3)} SinNeRF~\cite{xu2022sinnerf} adds a similar image patch regularizer by computing the feature differences by the DINO ViT~\cite{caron2021emerging}. Besides the semantic loss, SinNeRF also conducts an adversarial training strategy with the hinge GAN Loss~\cite{lim2017geometric}.
To integrate SinNeRF in our framework or panoramic NeRF learning, we randomly render projected image patches on these occluded regions and optimize the semantic loss extracted by the DINO ViT~\cite{caron2021emerging}. Besides the semantic loss, for SinNeRF we follow the adversarial training strategy that we train the NeRF model against a discriminator with the hinge GAN Loss~\cite{lim2017geometric}.

{\bf 4)} Omni-NeRF~\cite{gu2022omni} synthesizes novel panorama views from a single RGBD panorama. We use our predicted depth map as its depth input. Omni-NeRF is trained by multiple "pseudo ground truth" RGB panorama images of the sampled poses by point cloud projection. 

{\bf 5)} 360FusionNeRF~\cite{kulkarni2022360fusionnerf} also synthesizes novel panorama views from a single RGBD panorama. It also adds semantic consistency regularization by the CLIP~\cite{radford2021learning} loss but on the whole panorama instead of perspective views. The main difference to DietNeRF is that it directly computes CLIP loss on the whole rendered panorama images instead of the perspectively projected images.
\begin{table}[t]
\caption{Quantitative Comparison on the Replica dataset.}\label{tab:Comparisons_replica}
\vspace{-15pt}
\begin{center}
\scalebox{0.9}{
\begin{tabular}{c|c|c|c|c}
\toprule
~&PSNR$_{\rm mask.}$$\uparrow$&PSNR $\uparrow$&SSIM$\uparrow$&LPIPS$\downarrow$\\
\hline
\hline
DS-NeRF 
& 11.73
& 23.29
& 0.834 
& 0.265
\\
DietNeRF 
& 11.52
& 23.24 
& 0.836   
& 0.291
\\
SinNeRF 
& 9.80
& 22.70
& 0.826
& 0.251
\\
360FusionNeRF 
& 11.87
& 23.20
& 0.833
& 0.245
\\
Omni-NeRF 
& 10.42
& 18.98
& 0.790
& 0.522
\\
PERF
& \textbf{12.63}
& \textbf{23.49}  
& \textbf{0.838}
& \textbf{0.244}
\\
\bottomrule
\end{tabular}}
\end{center}
\vspace{-0.5cm}
\end{table}

\subsection{Comparisons on the Replica Dataset}\label{subsec:exp_replica}

\begin{figure*}[t]
  \centering
\includegraphics[width=0.9\linewidth]{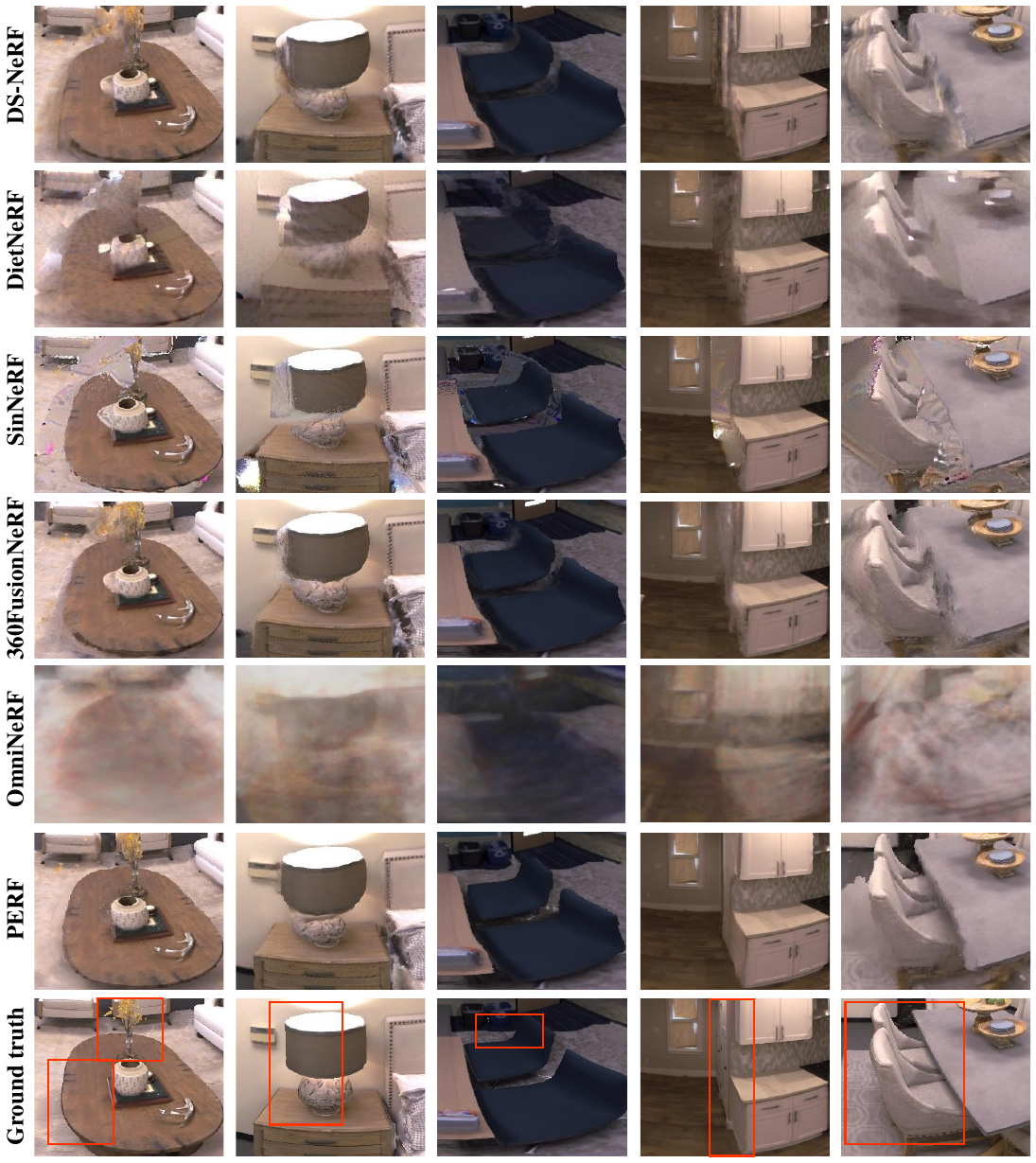}
  \caption{Visual comparison on the Replica dataset. One can observe foggy artifacts in the occluded regions of DS-NeRF, which is due to the incapability of a pure NeRF model to infer occluded content semantically. The renderings of DietNeRF and SinNeRF do not follow the coherent semantic contexts of the scene. In contrast, PERF synthesizes coherent novel views and has a smooth transition to the occluded region.}
  \label{fig:replica}
  \vspace{-10pt}
\end{figure*}

We compare PERF with the state-of-the-art NeRF methods on the Replica dataset.
The quantitative result in the Replica Dataset is shown in Table~\ref{tab:Comparisons_replica}. Not surprisingly, PERF achieves the best performance in terms of all the metrics. More importantly, our method achieves significant improvement on masked PSNR, which demonstrates that our method can well-inpaint invisible regions compared with other state-of-the-art methods. In Fig.~\ref{fig:replica}, we provide more visual comparisons. As the figure shows, several foggy effects in the occluded regions appear in the rendered results of DS-NeRF. This is because the NeRF model itself does not have the ability to semantically infer the content of the occluded regions without extra supervision.
The rendered results of DietNeRF, SinNeRF, and 360FusionNeRF are of better fidelity than those of DS-NeRF, without foggy effects.
However, the contents filled in the occluded regions do not well follow the semantic contexts, which looks unnatural for the human sense. 
The synthesized results of OmniNeRF are blurred because only the color information is used to supervise the NeRF model without geometric regularization.
Thanks to the strong inpainting ability of the Stable Diffusion model, our method PERF synthesizes the images with the best visual quality and the inferred contents in the occluded regions can smoothly transition to the visible regions of the reference view.

\begin{figure*}[t]
  \centering
  \includegraphics[width=0.9\linewidth]{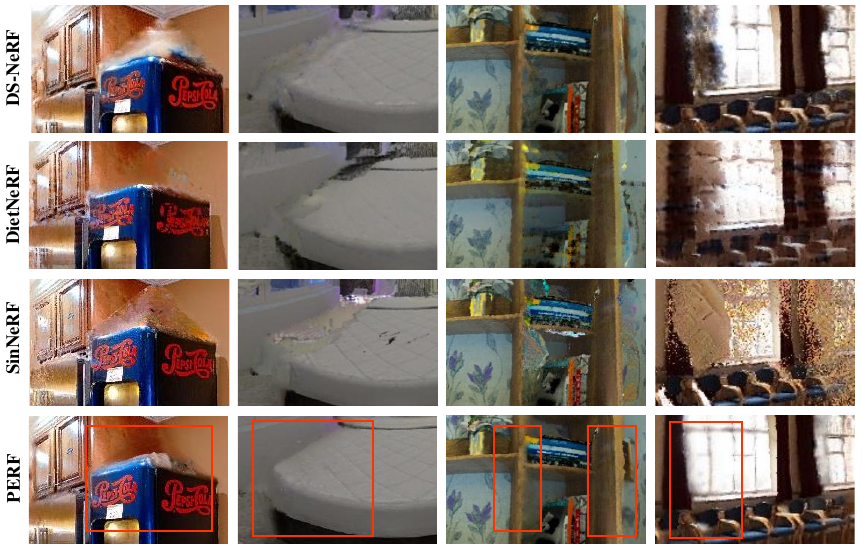}
  \vspace{-0pt}
  \caption{Visual comparison on the PERF-in-the-wild dataset. In the more challenging real-world scenarios, PERF significantly outperforms other methods in terms of visual quality and semantic coherence. For example, reasonable structures of windows (bottom row) and smooth transition of the occluded corners (top row).}
  \label{fig:SinPanoNVS-in-the-wild}
  \vspace{-0pt}
\end{figure*}

\subsection{Comparisons on the PERF-in-the-wild Dataset}
\label{subsec:perf-wild}
\textbf{Qualitative comparison.}
To demonstrate the robustness of PERF, we additionally evaluate PERF and the baseline few-shot NeRF methods on our collected in-the-wild dataset. Since there is no ground truth geometry for the in-the-wild images, we only show qualitative comparisons of the novel views. Fig.~\ref{fig:SinPanoNVS-in-the-wild} shows the visual comparisons, where the synthesis results of PERF are of the best visual quality.

\begin{figure}[h]
  \centering
  \includegraphics[width=0.8\linewidth]{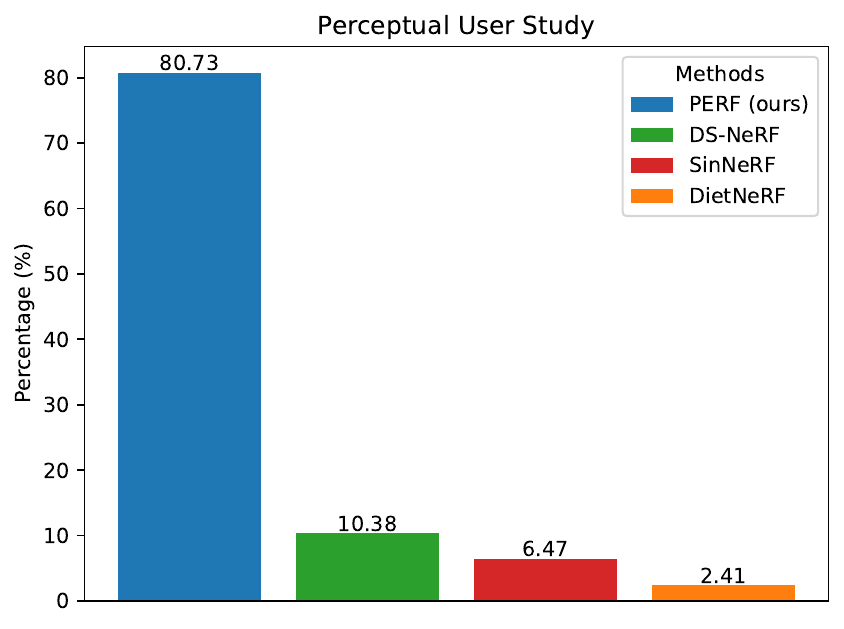}
  \vspace{0pt}
  \caption{User study on PERF-in-the-wild dataset. PERF is the most preferred among all the methods including DS-NeRF, SinNeRF, and DietNeRF. The experimental results demonstrate the effectiveness of our proposed method (Best viewed in color).}
  \label{fig:user_study}
  \vspace{0pt}
\end{figure}

\begin{figure}[h]
  \centering
  \includegraphics[width=1.0\linewidth]{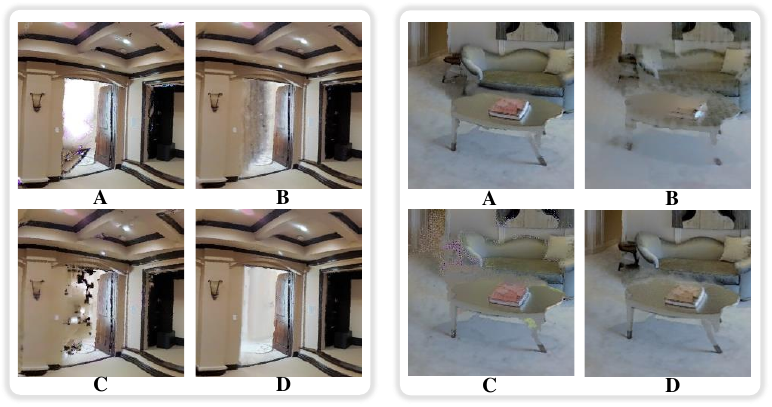}
    \caption{Two example questions of a user study on PERF-in-the-wild dataset. For these two questions, the options ``D'' (left) and ``A'' (right) represent PERF's results.} 
  \label{fig:user_study_example} 
\end{figure}

\noindent\textbf{User Study.} 
Since there is no ground truth geometry for the in-the-wild images, we only show qualitative comparisons of the novel views on the PERF-in-the-wild dataset. To complement the lack of quantitative results on PERF-in-the-wild, we conduct a user study to compare the visual effects of various single-view methods. We render the novel panorama images predicted by three methods~\cite{xu2022sinnerf,jain2021putting,deng2022depth} and ours, and show the cropped regions that are invisible from the input views. For each question, users were asked to pick the image that best meets the following criteria in their mind: 1)~Clearness - The image should not have any blurring, noise, or other visual distortions. 2)~Semantic coherence - The image should not have any jarring or unrelated elements that do not fit with the overall theme or meaning of the image. {Fig. \ref{fig:user_study_example}} shows two example questions of the user study. We rendered 16 examples, and 29 users participated in the study. Results are shown in {Fig. \ref{fig:user_study}}. We can observe that about 81\% prefer our rendered results, which significantly outperforms other methods.

\subsection{Ablation Study}

\begin{table}[t]
\caption{Ablation Studies. Without the proposed inpainting-and-erasing technique, our method shows degraded performance due to the geometry conflicts (Fig.~\ref{fig:PERF-vis-ablation}). This is because our method inpaints good semantics from a view and erases conflicted regions from other views.}\label{tab:ablation}
\begin{center}
\begin{tabular}{c|c|c}
\toprule
& w/o inpainting-and-erasing & Ours\\
\hline
PSNR $\uparrow$ & 23.34 & {\bf 23.49} \\
SSIM $\uparrow$ & 0.837 & {\bf 0.838} \\
LPIPS $\downarrow$ & 0.246 & {\bf 0.244} \\
PSNR$_{\rm mask.}$$\uparrow$ & 12.23 & {\bf 12.63} \\
\bottomrule
\end{tabular}
\end{center}
\end{table}

\begin{figure}[t]
  \centering
  \includegraphics[width=1.0\linewidth]{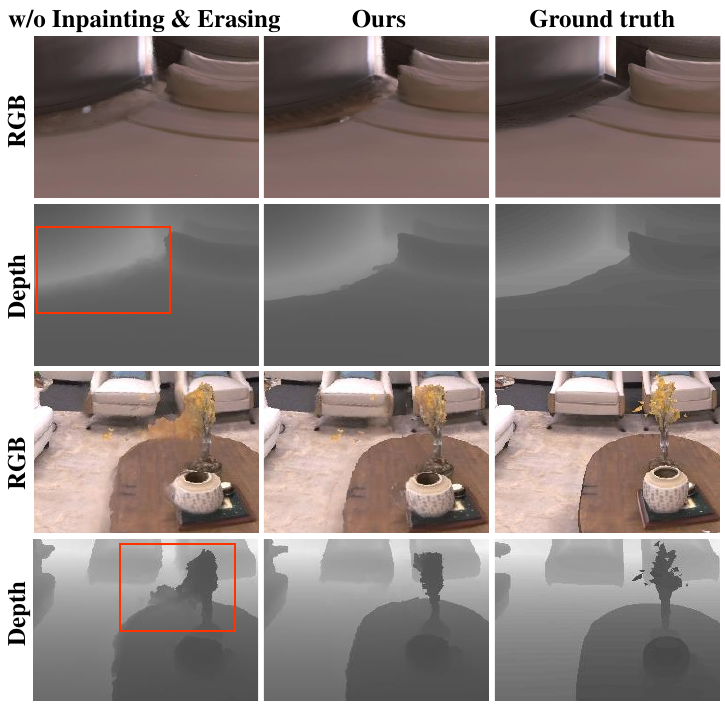}
  \vspace{0pt}
  \caption{The effect of the inpainting-and-erasing strategy. Without inpainting-and-erasing, the wrong geometry could appear and hurt the quality of the synthesized views as well.}
  \label{fig:PERF-vis-ablation}
  \vspace{0pt}
\end{figure}

We conduct the ablation study on the Replica dataset. In this experiment, we evaluate PERF without our proposed inpainting-and-erasing strategy, i.e., we do not check geometry conflicts and let all the inpainted results participate in the supervision of the NeRF model. As shown in Table~\ref{tab:ablation}, our method without the inpainting-and-erasing strategy shows worse quantitative results in terms of all the metrics. We further provide the visual results in Fig.~\ref{fig:PERF-vis-ablation}, where PERF with the full method designs has better quality in both the rendered images and the depths. Without our inpainting-and-erasing strategy, wrong redundant geometry appears, consequently hurting the synthesized RGB values. The experimental results demonstrate that the inpainting-and-erasing method is able to avoid conflicted geometry from different views, and thus achieves better 3D scene generation. 

\subsection{Applications}
\begin{figure}[t]
  \centering
  \includegraphics[width=1.0\linewidth]{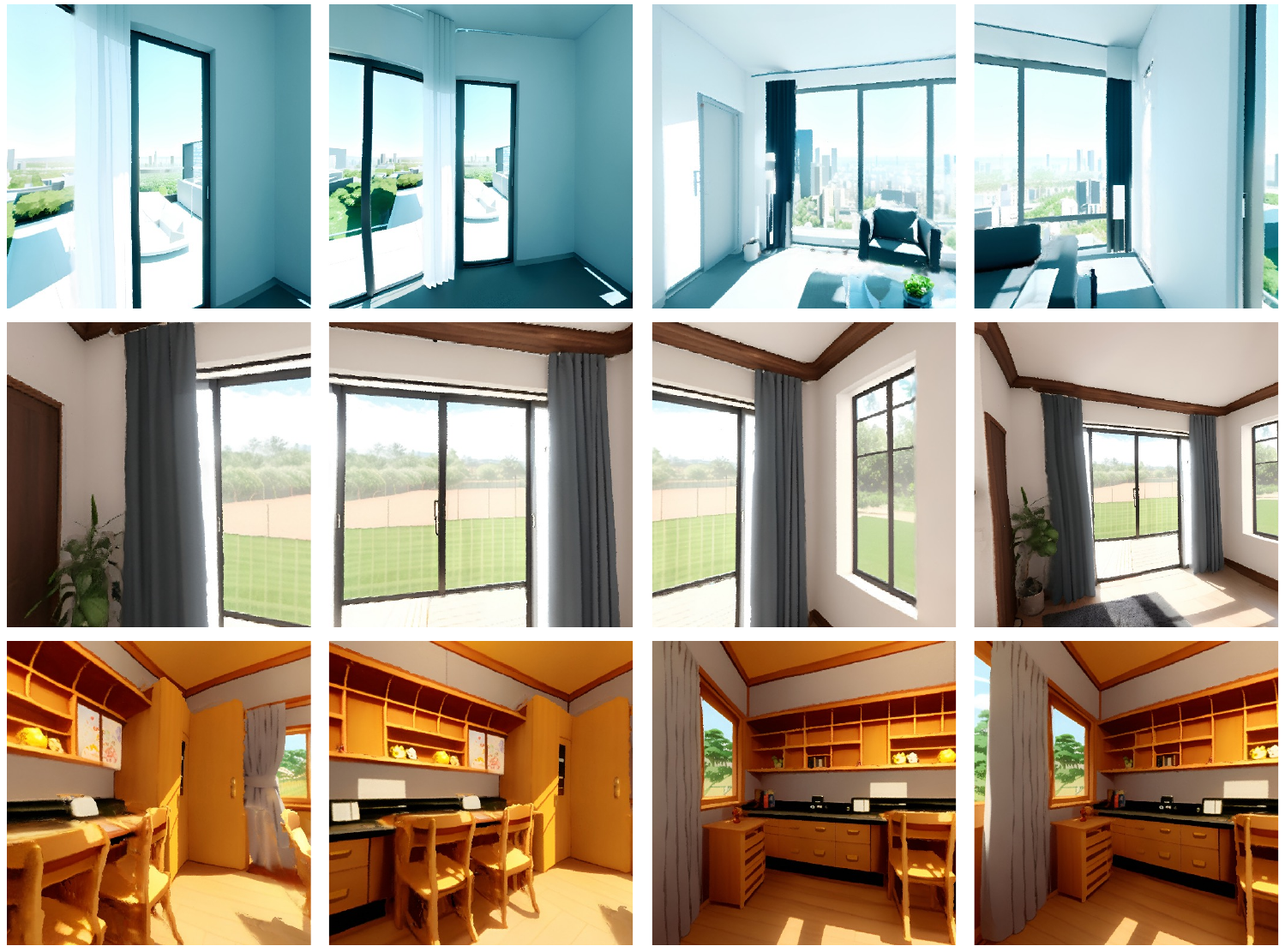}
  \vspace{0pt}
  \caption{{\bf Text-to-3D}. Each row shows an example. The sampled frames are presented from left to right by time. We use Blockade Labs sky box~\cite{blockade2023} to generate the panorama and we produce the 3D roaming effect by our PERF. Please refer to the project page for the text and the rendered videos.}
  \label{fig:text_to_3D_examples}
  \vspace{0pt}
\end{figure}

\begin{figure}[t]
  \centering
  \includegraphics[width=1.0\linewidth]{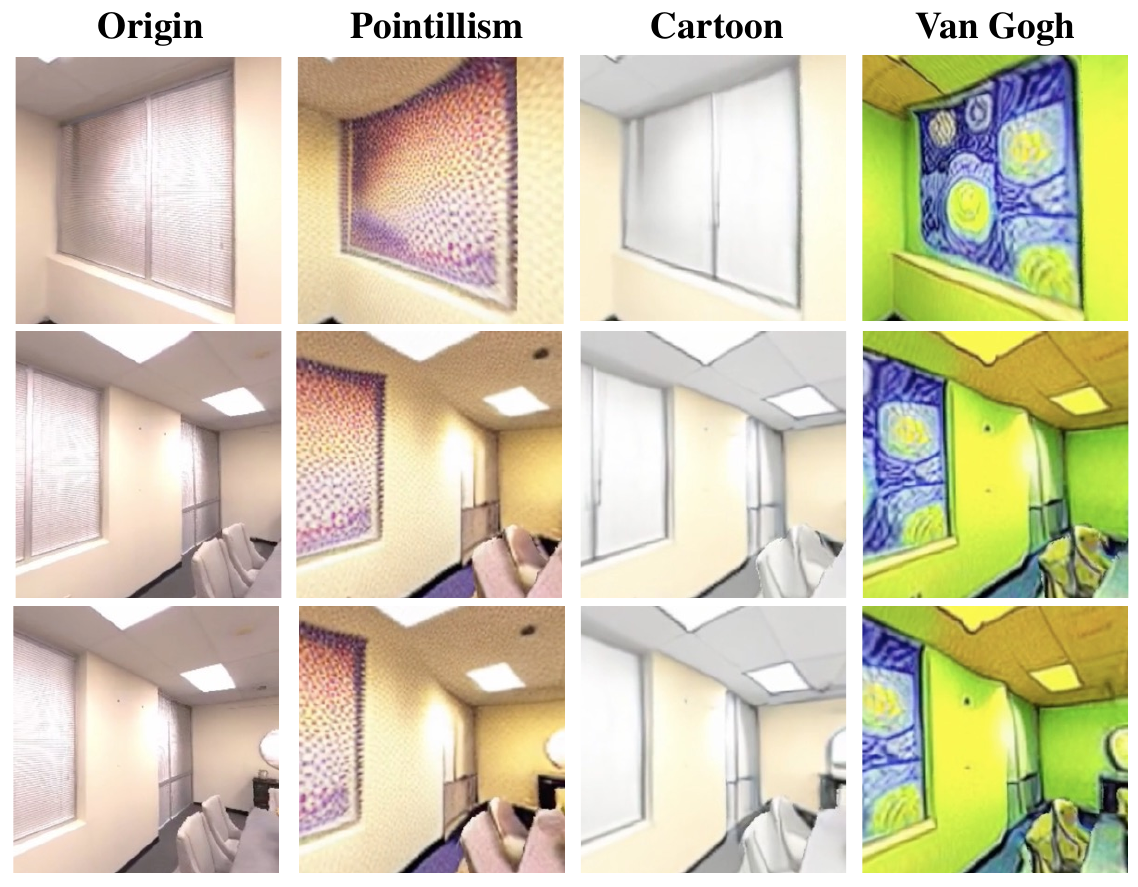}
  \vspace{0pt}
  \caption{{\bf 3D Scene Stylization}. We use InstructPix2Pix~\cite{brooks2022instructpix2pix} to change the 2D panorama image style and we can synthesize a 3D-consistent stylized scene via PERF. Please refer to the project page for rendered videos.}
  \label{fig:3d_stylization_examples}
  \vspace{0pt}
\end{figure}

Our PERF is originally designed for single panorama novel view synthesis, which can be also extended to some other applications like {\bf text-to-3D}, and {\bf 3D scene stylization}. For these extended applications, we combine related technologies to achieve these applications, as briefly shown in Fig. \ref{fig:text_to_3D_examples} and Fig. \ref{fig:3d_stylization_examples}. We invite the readers to watch detailed rendered videos on \textbf{our project page}. 

\textbf{Panorama-to-3D.} PERF enables panoramic indoor scene reconstruction from a single panorama, which can be observed in Figs. \ref{fig:replica} and \ref{fig:SinPanoNVS-in-the-wild}. PERF has a wide range of applications in real-world scenes, such as 3D virtual tours, virtual-reality games/conferences, and filmmaking.

\textbf{Text-to-3D.} Current text-to-3D methods \cite{raj2023dreambooth3d,metzer2022latent,poole2022dreamfusion,seo2023let} mainly focus on object-centric 3D generation driven by a text. We find that combing PERF and text-driven 2D panorama generation methods, such as Blockade Labs \footnote{\url{https://skybox.blockadelabs.com/}} or Text2Light \cite{chen2022text2light}, can generate promising 3D scenes. We first generate a 2D panorama with a text, and then reconstruct a 3D scene with the 2D panorama. The results are shown in Fig. \ref{fig:text_to_3D_examples}.

\textbf{3D Scene Stylization.} Since PERF is a method that reconstructs a 3D scene from a single panorama, it is easy to stylize the 3D scene. Given a panorama $I$, we can reconstruct a 3D scene $S$ by PERF. If we want to stylize the 3D scene $S$ to a new 3D scene $S^{'}$, we only need to stylize the 2D panorama $I$ to $I^{'}$ as a reference view and then reconstruct $S^{'}$ by PERF. We use InstructPix2Pix \cite{brooks2022instructpix2pix} to perform style transfer of 2D panoramas and synthesize stylized 3D scenes by PERF. The results are shown in Fig. \ref{fig:3d_stylization_examples}.

\section{Conclusion}
In this paper, we introduce a new method to train a panoramic NeRF from a single panorama. We present PERF, a 360-degree novel view synthesis framework that trains a panoramic neural radiance field from a single panorama. Specifically, given a single panorama, we first predict an initial depth map with a depth estimator and reconstruct visible 3D geometry with volume rendering. Then we integrate a collaborative RGBD inpainting approach into a NeRF to complete the invisible regions of RGB images and depth maps from random views. Moreover, we introduce a progressive inpainting-and-erasing generation method to avoid inconsistent geometry between visible regions and inpainted invisible regions. Extensive experiments on Replica and in-the-wild scenes demonstrate the superiority of our PERF over state-of-the-art methods. 

\noindent\textbf{Limitation.} PERF significantly improves the performance of single-shot NeRF, but heavily depends on the accuracy of the depth estimator and the Stable Diffusion.



\ifCLASSOPTIONcompsoc
  \section*{Acknowledgments}
\else
  \section*{Acknowledgment}
\fi

This work is supported by the National Research Foundation, Singapore under its AI Singapore Programme (AISG Award No: AISG2-PhD-2021-08-019), NTU NAP, MOE AcRF Tier 2 (T2EP20221-0012), and under the RIE2020 Industry Alignment Fund - Industry Collaboration Projects (IAF-ICP) Funding Initiative, as well as cash and in-kind contribution from the industry partner(s).

\ifCLASSOPTIONcaptionsoff
  \newpage
\fi



%
\bibliographystyle{IEEEtran}
\bibliography{egbib}

%








\end{document}